# Filling gaps in trustworthy development of AI

Incident sharing, auditing, and other concrete mechanisms could help verify the trustworthiness of actors

By Shahar Avin[1*†], Haydn Belfield[1,2], Miles Brundage[3], Gretchen Krueger[3], Jasmine Wang[4], Adrian Weller[2,6,7], Markus Anderljung[8], Igor Krawczuk[9], David Krueger[5,6], Jonathan Lebensold[4,5], Tegan Maharaj[5,10], Noa Zilberman[11]

The range of application of artificial intelligence (AI) is vast, as is the potential for harm. Growing awareness of potential risks from AI systems has spurred action to address those risks, while eroding confidence in AI systems and the organizations that develop them. A 2019 study (1) found over 80 organizations that published and adopted "AI ethics principles'', and more have joined since. But the principles often leave a gap between the "what" and the "how" of trustworthy AI development. Such gaps have enabled questionable or ethically dubious behavior, which casts doubts on the trustworthiness of specific organizations, and the field more broadly. There is thus an urgent need for concrete methods that both enable AI developers to prevent harm and allow them to demonstrate their trustworthiness through verifiable behavior. Below, we explore mechanisms (drawn from (2)) for creating an ecosystem where AI developers can earn trust - if they are trustworthy. Better assessment of developer trustworthiness could inform user choice, employee actions, investment decisions, legal recourse, and emerging governance regimes.

Common themes in statements of AI ethics principles include (i) assurance of safety and security of AI systems throughout their lifecycles, especially in safety-critical domains; (ii) prevention of misuse; (iii) protection of user privacy and source data; (iv) ensuring systems are fair and minimize bias, especially when such biases amplify existing discrimination and inequality; (v) ensuring the decisions made by AI systems, as well as any failures, are interpretable, explainable, reproducible and allow challenge or remedy; and (vi) identifying individuals or institutions who can be held accountable for the behaviors of AI systems. These principles address concerns that include accidents in robotic systems; erroneous judgements from AI systems used by physicians or in court; misuse of AI in surveillance, manipulation, or warfare; and risks to privacy and concerns about systemic bias (3).

In the study of trust in technology, a common approach differentiates trust in people (individuals and institutions) and trust in technology artifacts (4). While trust in artifacts mainly relies on competence and reliability, trust in people also relies on motives and integrity. Trust can be earned by providing reliable evidence that AI systems, and the processes used to develop and deploy them, address potential harms. This evidence carries further weight in an ecosystem where principles for preventing harms are well established, and where failure to adhere to principles carries meaningful consequences; a failure to establish an ecosystem that links trust to trustworthiness could spread into a general loss of trust in AI systems, compounding the harm from specific systems with the harm of foregone benefits. Concerns regarding motives, while crucial to some aspects of trust, are mostly outside the scope of the proposed mechanisms.

Trustworthy AI development presents significant challenges. Technical standards that assure an AI system adheres to the ethical principles mentioned are often lacking. Thus, experts need to evaluate specific AI systems in the contexts where they are developed and deployed. Experts may not be incentivized to address potential harms from their own organizations, and cooperation across organizations can raise antitrust law concerns. The proposed mechanisms we describe help address these challenges by sharing relevant information and incentivizing expert evaluation, which together can inform public assessments of AI developers' trustworthiness (Figure 1).

Beyond AI development, we recognize that the broader socio-technical context, including but not limited to AI procurement, deployment, social context, and use, will require additional engagement and measures. While our mechanisms focus on AI systems at or close to deployment, where requirements and context are clearer, they also extend to earlier stages of development. Separately, we note the need for AI developers to earn trust by consistently displaying trustworthy behavior more generally, including healthy, equitable, and diverse work environments, clear anti-retaliation policies to protect whistle-blowers, and broad environmental, ethical, and social responsibilities.

## MECHANISMS

### Red team exercises

To address concerns of misuse and novel vulnerabilities, a growing number of AI developers are turning to "red teams": professionals who consider a system from the perspective of an adversary, to identify exploitable vulnerabilities, which can then be mitigated. To date, AI red teams exist mostly within large industry and government labs, though experts also engage in "red team" activity in academia and through consultancy. AI red-teaming could form a natural extension of the cybersecurity red team community, though the data-driven and increasingly general nature of AI systems requires new domains of expertise.

We see space for the formation of a community of AI red team experts that shares experience across organizations and domains. Such exchange is not currently commonplace, though there has been a growing trend to publish threat modelling of AI systems (5). For example, there are public technical discussions on the feasibility of criminals using adversarial attacks on ML models, or on the possibility of misusing large-scale language models for online disinformation. Red-teaming could be carried out by an independent third party to address antitrust concerns (*6*).

[1]Centre for the Study of Existential Risk, University of Cambridge, Cambridge, UK. [2]Leverhulme Centre for the Future of Intelligence, University of Cambridge, Cambridge, UK. [3]OpenAI, San Francisco, USA. [4]McGill University, Montreal, Canada. [5]Mila, Montreal, Canada. [6]Department of Engineering, University of Cambridge, Cambridge, UK. [7]The Alan Turing Institute, London, UK. [8]Centre for the Governance of AI, Oxford, UK. [9]École Polytechnique Fédérale de Lausanne, Lausanne, Switzerland. [10]Faculty of Information, University of Toronto, Canada. [11]Department of Engineering Science, University of Oxford, Oxford, UK. Email: sa478@cam.ac.uk
†Summarizes and updates a 59-author report (2). After the corresponding author, the next five authors (2-6) and the last 6 authors (7-12) each form sets; within each set authors contributed equally and are ordered alphabetically.

### *Audit trails*

External audits, whether mandated by regulation or undertaken voluntarily, would form an important piece of the AI trustworthiness ecosystem (see below). To enable audits, AI developers would need to adopt best practices for documenting their development process and systems' makeup and activities. Clear standards for retaining information during development and operations exist in other domains (*7*). Standards and logging mechanisms have yet to be created to cover the range of AI applications, though there are some ongoing domain-specific efforts, e.g. for automated vehicles (*8*). While industry standards can raise antitrust concerns, lessons from other safety-critical industries suggest such concerns can be addressed if standards are mandated by governments, or if they are voluntary, developed in an open and participatory manner, or if accessible on fair, reasonable, and non-discriminatory terms (*6*).

Early progress can be seen in ethics frameworks that formalize questions to ask during the development process (e.g., Rolls Royce Aletheia Framework or the Machine Intelligence Garage Ethics Framework), in emerging guidelines for documenting certain features of AI models (e.g., ABOUT ML or Model Cards), and in proposals for continuous monitoring and logging. Developers are expected, for example, to record the provenance of all data used to train models, and to record outcomes of benefit and risk assessments conducted prior to deployment. Further progress requires collective effort to develop widely accessible and free standards for audit trails.

### *Interpretability and explainability*

Assuring the safety, accountability and fairness of AI systems is often challenged by their "black box" nature. Many researchers aim to address this challenge, either by restricting AI systems to human-readable, rules-based behaviors, or by explaining systems' outputs, e.g. by highlighting salient inputs. Research in this area has highlighted the importance of specific principles: (i) methods should provide sufficient insight for the end-user to understand how a model produced an output, (ii) the interpretable explanation should be faithful to the model, accurately reflecting its underlying behavior, and (iii) techniques should not be misused by AI developers to provide misleading explanations for system behavior. The challenge remains to translate these principles into verifiable practice.

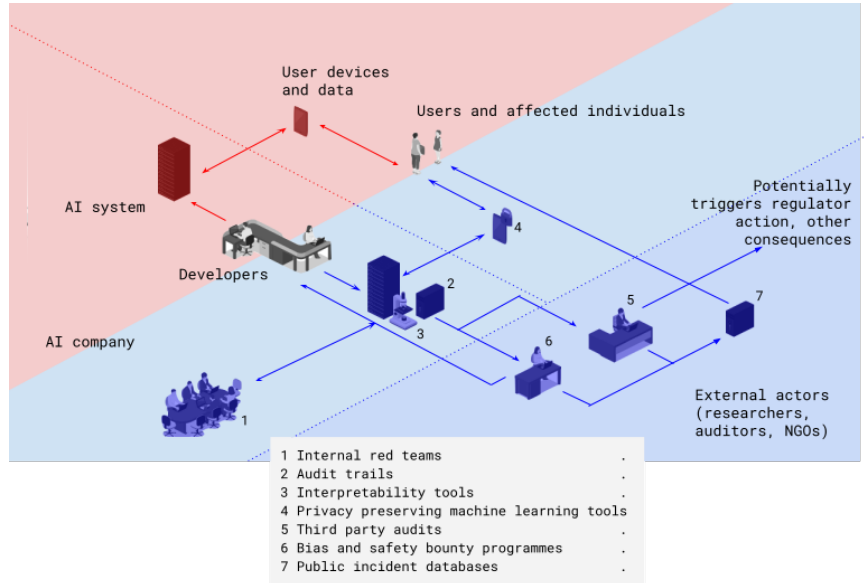

*Figure 1.* Existing relations (red) between organizations developing AI and users often leave gaps that make users unable to verify the trustworthiness of these organizations. The proposed relations (blue) are supported by the mechanisms described in the text, which either (i) help developers adopt best practices in their internal processes and handling of user data (1-4, light blue background), or (ii) incentivize external actors to evaluate the trustworthiness of developers and systems (5-7, dark blue background) and share that information with users. Together, these mechanisms promote a flow of information about trustworthiness, from developers, via external actors, to users.

Growth in interpretability and explainability research, for example in attribution methods that reliably explain specific predictions of computer vision models, is welcome. However, more research effort should be directed towards combining techniques to address ethical and safety concerns. For example, greater effort could be focused on increasing the explainability of accidents and systemic biases. Further, we should (i) develop common standards for domain-specific interpretability criteria and objectives, e.g. that can faithfully explain possible differences in predictions for different populations, (ii) develop tests that measure compliance with such standards, and (iii) identify domains where the context of use requires a high level of interpretability, and develop performant interpretable models for these domains.

Complementary to interpretability and explainability, which help interrogate the outputs of AI systems, reproducibility allows external teams to recreate an AI system and interrogate it, verifying claims made by developers. Initiatives like the ACM artifact review and badging and the ML reproducibility challenge incentivize reproducibility in research settings.

### *Privacy-preserving machine learning*

Concerns regarding privacy in ML include unauthorized access to the data used to train models, privacy violations and targeting of individuals and communities through inferring sensitive information from trained models, and unauthorized access to the trained model itself. Research on "privacy-preserving machine learning" (PPML) has developed complementary techniques to address these concerns. *Federated learning* techniques allow the centralized training of a model with decentralized data, without raw data ever leaving the source device (*9*). *Differential privacy* techniques modify the development process such that trained models retain meaningful statistical patterns at the population level but reduce the risk of inferring information about individuals (*10*). *Encrypted computation* allows data owners and model developers to train models without either side gaining access to the information of the other. Together these techniques can help mitigate privacy concerns, though each involves trade-offs, for example regarding ease of development or training efficiency.

PPML techniques still lack both standard software libraries and awareness amongst AI developers. However, a growing community of open-source implementation projects has enabled progress towards wider adoption. Projects that extend PPML to existing machine learning frameworks, such as PyTorch's Opacus and Tensorflow Privacy, have seen particular growth. Projects like FedAI, PySyft, Flower and OpenFL provide added value by considering privacy beyond the narrow context of model training; they illustrate to AI developers how to reason about where data is stored and what trade-offs exist in preserving privacy. To further accelerate adoption, we recommend establishing reliable support for active PPML projects, open standards, algorithm benchmarks and educational resources.

### *Third party auditing*

For best practices to engender trust, AI developers must follow them, and be seen to follow them. This is complicated by limitations on the information that can be shared publicly by AI developers, for example private user data. A solution adopted in several other industries is third party auditing, where an auditor gains access to restricted information and in turn either testifies to the veracity of claims made, or releases information in an anonymized or aggregated manner. Third-party auditing sidesteps several antitrust concerns (*6*); it is also well-positioned to leverage technical solutions, such as secure multiparty computation, that allow verification of claims without requiring direct access to sensitive information.

Auditing can take many forms, involving varying mixes of government and private actors and a range of funding models and information sharing practices. A recent concrete proposal for independent auditing of AI systems highlights three key ingredients: (i) proactive independent risk assessment, (ii) reliance on standardized audit trails, and (iii) independent assessment of adherence to guidelines and regulations (*11*).

Auditing can only contribute to trust if auditors themselves are trusted, and if failures to pass audits carry meaningful consequences; it is therefore essential that auditors have strong incentives to report their findings accurately and protections for raising concerns when necessary. Reputational mechanisms, as well as government and civil society backing, would help provide such incentives.

### *Bias and safety bounties*

The complexity of AI systems means it is possible for some vulnerabilities and risks to evade detection prior to release. For similar reasons, in the field of cybersecurity, experts research flaws and vulnerabilities in published software and publicly available hardware. What began as an antagonistic relationship between vendors and external security researchers led to the development of "bug bounties" and responsible disclosure: mechanisms by which security experts can carry out their research and be financially rewarded for their findings, while companies benefit from the discoveries and have a period in which to address them before they are publicly revealed. These mechanisms have helped align incentives in cybersecurity, have led to more secure systems, and helped increase trust in companies that meaningfully and continuously engage in bug bounty programs.

A similar approach could be adopted to reward external parties who research bias and safety vulnerabilities in released AI systems. At present, much of our knowledge about harms from AI comes from academic researchers and investigative journalists, who have limited access to the AI systems they investigate, and often experience antagonistic relationships with the developers whose harms they uncover. The Community Reporting of Algorithmic System Harms project from the Algorithmic Justice League explores the potential for bounty programs that cover a broad range of harms from AI systems, including unfair bias. This July, Twitter offered bounties to researchers who could identify biases in their image cropping algorithm.

Note that such bounty systems do not shift the burden from AI developers - more resources should also be invested in surfacing and addressing vulnerabilities and biases before product release.

### *Sharing of AI incidents*

As AI systems move from labs to the world, theoretical risks materialize in actual harms. Collecting and sharing evidence about such incidents can inform further R&D efforts as well as regulatory mechanisms and user trust. However, any individual AI developer is disincentivized to share incidents in their own systems, due to reputational harms, especially if they cannot trust competitors to share similar incidents - a classic collective action problem (*12*). Mechanisms are needed that enable coordination and incentivize sharing.

Incident sharing could become a regulatory requirement. Until then, voluntary sharing can be incentivized, for example by allowing anonymous disclosure to a trusted third party. Such a third party would need to have transparent processes for collecting and anonymizing information and operate an accessible and secure portal. The Partnership on AI is experimenting with such a platform through its AI Incident Database, where information about AI incidents is compiled both from public sources and reporting from developers (*13*). Recently, the Center for Security and Emerging Technologies developed a taxonomy of three categories (specification, robustness, and assurance) based on the incidents reported, with over 100 incidents exemplifying each category (*14*).

## Conclusion

The mechanisms outlined above provide concrete next steps that can help the public assess the trustworthiness of AI developers. While we stress the need for a broader ecosystem that considers stages both before and after development, and that can enforce meaningful consequences, we see the verification of trustworthy behavior by developers as an important part of the maturation of the field of AI. These mechanisms enable targeted and effective regulation; for example, the EU has proposed AI regulation that includes incident-sharing, audit trails, and third-party auditors (*15*). We invite greater engagement with this urgent challenge at the interface of interdisciplinary research and policy.

Acknowledgments: The report from which this paper draws was composed by 59 authors and benefitted from further feedback from colleagues and at workshops (see (2) for details). We also thank Sankalp Bhatnagar for graphic design assistance and two anonymous reviewers for insightful suggestions. S.A. received funding from the Isaac Newton Trust and from Templeton World Charity Foundation, Inc. The opinions expressed in this publication are those of the author(s) and do not necessarily reflect the views of Templeton World Charity Foundation, Inc. H.B. received funding from Casey and Family Foundation. M.B. and G.K. received funding from OpenAI. A.W. received funding from a Turing AI Fellowship under grant EP/V025379/1, The Alan Turing Institute, and the Leverhulme Trust via CFI. M.A. received funding from Open Philanthropy and support from the Future of Humanity Institute. I.K. acknowledges support from the ERC Horizon 2020 research and innovation programme under grant 725594 (time-data). D.K. received funding from Open Philanthropy.